\documentclass[twocolumn,headings=small,abstract=true,10pt]{scrartcl}

\usepackage[ngerman,english]{babel}
\usepackage[T1]{fontenc}
\usepackage[utf8]{inputenc}
\usepackage{mathptmx}
\usepackage[scaled]{helvet}
\usepackage[sort&compress,numbers]{natbib}
\usepackage{amsmath,amsfonts}
\usepackage{microtype}
\usepackage{overpic}
\usepackage[outline]{contour}
\usepackage[twocolumn,left=23mm,right=23mm,top=17mm,bottom=34mm]{geometry}

\setkomafont{sectioning}{\normalfont\bfseries}
\setkomafont{caption}{\small}
\setcapindent{1em}
\usepackage{tabu}

\usepackage{ntheorem}

\theoremstyle{plain}
\newtheorem{theorem}{Theorem}
\newtheorem{proposition}[theorem]{Proposition}

\theoremstyle{definition}
\theorembodyfont{\normalfont}

\theoremstyle{remark}
\theoremheaderfont{\itshape}
\theorembodyfont{\normalfont}
\newtheorem{remark}[theorem]{Remark}

\usepackage{algorithm}
\usepackage{algpseudocode}
\usepackage{todonotes}

\usepackage{hyperref}
\hypersetup{
  pdfauthor={Zijia Li, Tudor-Dan Rad, Josef Schicho, Hans-Peter Schröcker},  pdftitle={Factorisation of Rational Motions: A Survey with Examples and Applications},  pdfkeywords={motion polynomial, dual quaternions, mechanism synthesis, Bennett flip, multiplication trick},    hidelinks=true,}

\newcommand{\EG}{E}
\newcommand{\SQ}{S}
\newcommand{\R}{{\mathbb{R}}}
\renewcommand{\H}{{\mathbb{H}}}
\renewcommand{\DH}{{\mathbb{DH}}}
\newcommand{\cj}[1]{\overline{#1}}
\newcommand{\eps}{\varepsilon}
\newcommand{\qi}{{\mathbf{i}}}
\newcommand{\qj}{{\mathbf{j}}}
\newcommand{\qk}{{\mathbf{k}}}
\newcommand{\peq}{\equiv}
\newcommand{\SO}[1][3]{\mathrm{SO}(#1)}
\newcommand{\SE}[1][3]{\mathrm{SE}(#1)}
\providecommand{\newblock}{\relax}

\title{\Large Factorisation of Rational Motions:\\ A Survey with Examples and Applications}
\author{
  \normalsize
  \begin{tabu}{X[c,0.26]X[c,0.24]X[c,0.26]X[c,0.24]}
    Z.~Li\linebreak RICAM\linebreak Austrian Academy of Sciences\linebreak Linz, Austria &
    T.-D.~Rad\linebreak Unit Geometry and CAD\linebreak University of Innsbruck\linebreak Innsbruck, Austria&
    J.~Schicho\linebreak RISC\linebreak Johannes Kepler University\linebreak Linz, Austria &
    H.-P.~Schröcker\linebreak Unit Geometry and CAD\linebreak University of Innsbruck\linebreak Innsbruck, Austria
  \end{tabu}
                        }
\date{\normalsize\today}

\begin{document}

\twocolumn[\begin{@twocolumnfalse}

\maketitle
\begin{abstract}
  \par\noindent    Since its introduction in 2012, the factorization theory for
  rational motions quickly evolved and found applications in
  theoretical and applied mechanism science. We provide an accessible
  introduction to motion factorization with many examples, summarize
  recent developments and hint at some new applications. In
  particular, we provide pseudo-code for the generic factorization
  algorithm, demonstrate how to find a replacement linkage for a
  special case in the synthesis of Bennett mechanisms and, as an
  example of non-generic factorization, synthesize open chains for
  circular and elliptic translations.
 \end{abstract}
 \par\bigskip
\end{@twocolumnfalse}]

\section{Introduction}
\label{sec:introduction}

The factorization of rational motions was introduced in \cite{hegedues12:_factorization,  hegedus13:_factorization2} and has ever since proved its usefulness in mechanism synthesis. So
far, it has been used for the construction of new linkages \cite{hegedus13:_factorization2,  LS13,LS13CK,
  hegedus14:_four_pose_synthesis,lsharp,  li14:_spatial_straight_line_linkages,li15:_7R} with attractive properties for
applications, including rational parameterisations of all relative
motions in the linkage, computable (sometimes even linear)
input-output relations, and the availability of simple symbolic and
numeric synthesis algorithms.

The fundamental factorization theorem of
\cite{hegedus13:_factorization2} is a statement about ``generic''
rational motions. Current research focuses on non-generic cases.
Preliminary results are quite promising and will also be addressed in
this survey article. Related results can be found in \cite{li15:_7R}
(these proceedings).

We continue this paper with a brief introduction to dual quaternions
and motion polynomials in Section~\ref{sec:dual-quaternions}. It is
followed by algorithms and examples obtained from factorization in
generic cases (Section~\ref{sec:generic-factorization}). In
particular, we discuss a special case in the synthesis of Bennett
mechanisms where the two 2R dyads coincide and no valid 4R linkage can
be constructed. It is, however, possible to find a 5R or 6R linkage
that generates the same motion.

In Section~\ref{sec:exceptional-factorizations} we discuss circular
and elliptic translations as an exceptional case of motion
factorization. Here, the generic factorization algorithm fails but
straightforward computations give a result. By means of a recently
invented ``multiplication trick'' we can generate non-planar 4R chains
for elliptic translations and construct linkages from them.

\section{Dual Quaternions and Motion Polynomials}
\label{sec:dual-quaternions}
\vskip -0.2cm

Quaternions and dual quaternions already proved their usefulness in
computational kinematics and mechanism science. See
\cite{husty12:_kinematics_algebraic_geometry},
\cite[Ch.~4]{mccarthy90:_introduction_theoretical_kinematics} or
\cite[Ch.~9]{selig05:_geometric_fundamentals_robotics} for an
introduction. Here, we briefly summarize some basic definitions and
concepts that will be needed in later sections.

\subsection{Quaternions and dual quaternions}
\vskip -0.2cm

A quaternion can be written as $q = q_0 + q_1\qi + q_2\qj + q_3\qk$
with real numbers $q_0$, $q_1$, $q_2$, $q_3$. The set of all
quaternions is denoted by $\H$. Quaternions are added component-wise
and their multiplication is determined by the rules
\begin{equation*}
  \qi^2 = \qj^2 = \qk^2 = \qi\qj\qk = -1.
\end{equation*}
It is associative but not commutative. We have, for example
$(\qi + \qj)(2 - \qk) = 2\qi + 2\qj - \qi\qk - \qj\qk = 2\qi + 2\qj+\qj -
\qi = \qi + 3\qj$ but $(2 - \qk)(\qi + \qj) = 2\qi + 2\qj - \qk\qi -
\qk\qj = 2\qi + 2\qj - \qj + \qi = 3\qi + \qj$.

The conjugate quaternion to $q = q_0 + q_1\qi + q_2\qj + q_3\qk$ is
$\cj{q} = q_0 - q_1\qi - q_2\qj - q_3\qk$. Depending on the individual
author, the quaternion norm is either
$q\cj{q} = q_0^2 + q_1^2 + q_2^2 + q_3^2 \in \R$ or the square root of
$q\cj{q}$. We follow the first convention. Quaternions with unit norm
$q\cj{q} = 1$ can model rotations in $\SO$ via
\begin{equation}
  \label{eq:1}
  x = x_1\qi + x_2\qj + x_3\qk\ \mapsto\ y_1\qi + y_2\qj + y_3\qk = qx\cj{q}.
\end{equation}
Here, a vector $(x_1,x_2,x_3) \in \R^3$ is identified with the vector
quaternion $x_1\qi + x_2\qj + x_3\qk \in \H$. The oriented axis
direction of the displacement \eqref{eq:1} is $(q_1,q_2,q_3)$ and the
rotation angle $\varphi$ satisfies
$\tan(\varphi/2) = q_0(q_1^2+q_2^2+q_3^2)^{-1/2}$. Note that $q$ and
$-q$ describe the same rotation.

A dual quaternion can be written as $h = p + \eps q$ with $p$,
$q \in \H$. The \emph{dual unit} $\eps$ satisfies $\eps^2 = 0$ and
commutes with $\qi$, $\qj$, $\qk$. For example,
$(\qi + \eps \qj)(\qj - \eps\qk) = \qi\qj +\eps(\qj^2 - \qi\qk) +
\eps^2\qj\qk = \qk - \eps(1 - \qj)$.
The set of dual quaternions is denoted by $\DH$, the conjugate dual
quaternion is $\cj{h} = \cj{p} + \eps\cj{q}$, and the dual quaternion
norm is $h\cj{h} = p\cj{p} + \eps(p\cj{q} + q\cj{p})$. Because of
$p\cj{p}$, $p\cj{q}+q\cj{p} \in \R$, it is a dual number. An important
difference between quaternions and dual quaternions is that $h \neq 0$
does not imply existence of $h^{-1} \in \DH$ such that $hh^{-1} = 1$.
Dual quaternions different from the set $\{\eps d \mid d \in \H\}$ do
not possess such a multiplicative inverse. If $h$ is not from that
set, its multiplicative inverse is $h^{-1} = (h\cj{h})^{-1}\cj{h}$. It
equals the conjugate up to a scalar dual factor.

Unit dual quaternions model $\SE$, the group of rigid body
displacements. Identifying the point $(x_1,x_2,x_3)$ with the
dual quaternion $x = x_1\qi + x_2\qj + x_3\qk$, the action of
the unit dual quaternion $h = p + \eps q$ is
\begin{equation}
  \label{eq:2}
  x\ \mapsto\ px\cj{p}+p\cj{q} - q\cj{p} = px\cj{p}+2p\cj{q}.
\end{equation}
We see that the spherical component of the rigid body displacement is
described by the primal part $p$ while the translation is given by the
vector quaternion $p\cj{q}-q\cj{p} = 2p\cj{q}$.

Again, $h$ and $-h$ model the same rigid body displacement. This
suggests, to projectivise the space of dual quaternions by identifying
$h$ and $\lambda h$ for any $\lambda \in \R \setminus \{0\}$. This
turns $\DH$ into the real projective space $P^7$ of dimension
seven. All points (classes of proportional dual quaternions) that can
be normalized to unity describe rigid body displacements. These are
precisely the points of the quadric hypersurface $\SQ \subset P^7$
given by 
$p\cj{q} + q\cj{p} = 0$ (the \emph{Study quadric}) minus the
three-space given by $p = 0$ (the \emph{exceptional three space $\EG$}).

\subsection{Polynomials and Motion Polynomials}
\vskip -0.2cm

The set of polynomials in $t$ with coefficients in $\DH$ is denoted by
$\DH[t]$. Polynomial multiplication is defined by the convention that
the indeterminate $t$ commutes with all coefficients. For $h$,
$k \in \DH$ we have, for example
\begin{equation}
  \label{eq:3}
  C = (t - h)(t - k) = t^2 - (h + k)t + hk.
\end{equation}
The value of the polynomial $C = \sum_{i=0}^n c_it^i \in \DH[t]$ at
$h \in \DH$ is defined as $C(h) := \sum_{i=0}^n c_ih^i$. Note that
with this convention we have for the polynomial $C$ in \eqref{eq:3}
\begin{equation*}
  \begin{aligned}
  C(h) &= h^2 - (h + k)h + hk = hk - kh,\\
  C(k) &= k^2 - (h + k)k + hk = -hk + hk = 0.
  \end{aligned}
\end{equation*}
This seemingly strange behavior is a consequence of the definition
for multiplication and evaluation. Linear factors of polynomials over
$\DH$ are not directly related to zeros. There is, however, an
important exception (Lemma~2 of \cite{hegedus13:_factorization2}):

\begin{proposition}
  The polynomial $C \in \DH[t]$ can be factored as $C = C'(t-h)$ with
  $C' \in \DH[t]$ if and only if $C(h) = 0$.
\end{proposition}

In other words, the zeros of $C$ correspond to linear \emph{right
  factors} of~$C$.

The conjugate of the polynomial $C = P + \eps Q = \sum_{i=0}^n c_it^i$
is defined as $\cj{C} = \sum_{i=0}^n \cj{c_i}t^i$ and the norm
polynomial is $C\cj{C} = P\cj{P} + \eps(P\cj{Q} + Q\cj{P})$. Its
coefficients are dual numbers. A polynomial
$C = P + \eps Q \in \DH[t]$ is called a \emph{motion polynomial,} if
its leading coefficient is invertible and the norm polynomial is even
real. This important condition ensures that the curve parameterised by
$C(t)$ for $t \in \R$ lies in the Study quadric. Hence motion polynomials
act via \eqref{eq:2} on points $x = x_1\qi + x_2\qj + x_3\qk$ and
yield rational parametric equations
\begin{equation*}
  t \mapsto
  \frac{Px\cj{P} + P\cj{Q} - Q\cj{P}}{P\cj{P}} =
  \frac{Px\cj{P} + 2P\cj{Q}}{P\cj{P}},\quad t \in \R \cup \{\infty\}
\end{equation*}
for all trajectories. The extension of the parameter range to
$\R \cup \{\infty\}$ via $C(\infty) = c_n$ is necessary for obtaining
complete curves. Summarizing, we can say that \emph{motion polynomials
  parameterise rational motions.} Conversely, it follows from
\cite{juettler93:_rationale_bewegungsvorgaenge} that any motion with
only rational trajectories is given by a suitable motion polynomial.

Lets look at linear motion polynomials $C = c_1t + c_0$. Because $c_1$
is invertible, we may transform it to $C = t - h$ by left multiplying
with $c_1^{-1}$. This amounts to a change in the fixed coordinate
frame and is no loss of generality. The norm polynomial is
$(t - h)(t - \cj{h}) = t^2 - (h + \cj{h})t + h\cj{h}$ and the
conditions for a linear motion polynomial are
\begin{equation*}
  h + \cj{h} \in \R,\quad
  h\cj{h} \in \R.
\end{equation*}
The second condition implies that $h$ itself describes a rigid body
displacement, the first condition implies vanishing of the dual scalar
part. In coordinates we have
$h = h_0 + h_1\qi + h_2\qj + h_3\qk + \eps(h_5\qi + h_6\qj + h_7\qk)$
with $h_1h_5 + h_2h_6 + h_3h_7 = 0$. It is known that the rigid body
displacements $C(t)$, $t \in \R \cup \{\infty\}$ are either all
rotations about the fixed axis with Plücker coordinates
$[h_1,h_2,h_3,-h_5,-h_6,-h_7]$ or, if $h_1 = h_2 = h_3 = 0$,
translations in direction $(h_5,h_6,h_7)$.

\begin{remark}
  \label{rem:angle}
  In case of a rotation, the rotation angle $\varphi$ satisfies
  $\tan(\varphi/2) = (t-h_0)(h_1^2+h_2^2+h_3^2)^{-1/2}$. Using the
  coefficients $a = -h-\cj{h}$ and $b=h\cj{h}$ of the norm polynomial,
  this becomes $\tan(\varphi/2) = (2t+a)(4b-a^2)^{-1/2}$. \emph{The
    rotation angle at the parameter value $t$ is hence already
    determined by the norm polynomial of $t-h$.}  A similar statement
  holds true for the translation distance in case of
  $h_1 = h_2 = h_3 = 0$. Also note that $h$ is a zero of the norm
  polynomial $(t-h)(t-\cj{h})$.
\end{remark}

\section{Generic Factorization}
\label{sec:generic-factorization}
\vskip -0.2cm

A motion polynomial $C = P + \eps D$ is called \emph{generic,} if its
primal part $P$ has no real factors. If $P$ has a real factor with
(possibly complex) zero $t_0$, the point $C(t_0)$ lies in the
exceptional generator $\EG$ of the \emph{Study quadric} $\SQ$. Hence, $C$ is
not generic, if it intersects $\EG$ (possibly over the complex
numbers) or, equivalently, if the degree of the spherical motion
component, which is parameterised by $P$, can be reduced by dividing
off a real factor. The fundamental theorem in motion factorization is

\begin{proposition}[\cite{hegedus13:_factorization2}, Theorem~1]
  \label{prop:factorization}
  If $C$ is a monic, generic motion polynomial of degree $n$, there
  exist rotation polynomials $t-h_1$, \ldots, $t-h_n$ such that
  \begin{equation}
    \label{eq:4}
    C = (t-h_1) \cdots (t-h_n).
  \end{equation}
\end{proposition}

We call \eqref{eq:4} a factorization of $C$. If $C$ is not generic,
factorizations of the shape \eqref{eq:4} may or may not exist. It is
also noteworthy and very important for applications in mechanism
science that the factorization \eqref{eq:4} is, in general, not unique
(compare Algorithm~\ref{alg:factorization} below). It can be viewed as
a decomposition of a rational motion into the product of rotations,
whose angles are linked via the common parameter $t$. In other words,
every factorization of the motion polynomial $C$ corresponds to an
open $n$R chain whose end-effector can follow the motion parameterised
by $C$. Combining several of these open chains will in general result
in an over-constrained linkage with end-effector motion parameterised
by~$C$.

\subsection{Computation}
\vskip -0.2cm

Computing a factorization \eqref{eq:4} of a generic motion polynomial
$C = P + \eps Q$ is not much harder than the factorization of real
polynomials over the complex numbers. An important ingredient is right
division for polynomials in $\DH[t]$. Given $A$ and $B \in \DH[t]$
such that the leading coefficient of $B$ is invertible, there exist
unique polynomials $Q$ and $R$, called \emph{quotient} and
\emph{remainder}, such that $A = QB + R$ and $\deg R < \deg B$.
Algorithm~\ref{alg:qr} computes $Q$ and $R$ under the additional
assumption that $B$ is monic. This is no real loss of generality and
is usually fulfilled in our applications. Note that there is no
essential difference between Algorithm~\ref{alg:qr} and a standard
algorithm for polynomial division in $\R[t]$. In particular, it does
not require any quaternion division. Because of non-commutativity it
is, however, important to strictly adhere to the given order of
factors (for example, $Bc$ in Line~\ref{li:Bc} must not be replaced by
$cB$).

\begin{algorithm}
  \caption{Polynomial division}
  \label{alg:qr}
  \begin{algorithmic}[1]
    \Require $A,B \in \DH[t]$, $B$ monic.
    \Ensure $Q,R$ such that $A = QB + R$ and $\deg R < \deg B$.
    \Statex
    \State $Q \gets 0$, $R \gets A$
    \While{$\deg(R) \ge \deg(B)$}
      \State $c \gets \text{leading coefficient of $R$}$
      \State $n \gets \deg(R) - \deg(B)$
      \State $Q \gets Q + c t^n$
      \State $R \gets R - B c t^n$ \label{li:Bc}
    \EndWhile
    \State\Return $Q,R$
  \end{algorithmic}
\end{algorithm}

Algorithm~\ref{alg:factorization} can be used to compute a
factorization of a generic, monic motion polynomial $C$. (If the
leading coefficient $c_n$ of $C$ is different from one, we can compute
a factorization of $c_n^{-1}C$ and left-multiply it with $c_n$.) By
genericity, the norm polynomial $C\cj{C}$ has no real zeros. Hence,
there exist irreducible quadratic polynomials
$M_1,\ldots,M_n \in \R[t]$ such that $C\cj{C} = M_1 \cdots M_n$. The
factorizations of $C$ are labeled by the permutations of these
factors. The functions $\Call{Remainder}{C,M_n}$ in
Line~\ref{li:remainder} and $\Call{Quotient}{C, t-h_i}$ in
Line~\ref{li:quotient} are derived from Algorithm~\ref{alg:qr}. In
Line~\ref{li:zero} we have to compute the zero $r_1^{-1}r_0$ of a
linear polynomial $R = r_1t + r_0$. It exists and is unique if $r_1$
is invertible. Again, this is guaranteed by genericity of~$C$.

From Algorithm~\ref{alg:factorization} we infer further properties of
motion factorization. They follow from the labeling of factorizations
by quadratic factors of $C\cj{C}$ and Remark~\ref{rem:angle}.

\begin{proposition}
  In general, a motion polynomial of degree $n$ admits $n!$ different
  factorizations. In any factorization, the factors labeled by the
  same quadratic factor $M_i$ of the norm polynomial give rise to
  rotations with the same angular velocity.
\end{proposition}

\begin{algorithm}
  \caption{Factorization of motion polynomials}
  \label{alg:factorization}
  \begin{algorithmic}[1]
    \Require Generic, monic motion polynomial $C \in \DH[t]$ of degree
             $n$, tuple $F = (M_1,\ldots,M_n)$ of quadratic,
             irreducible real factors of $C\cj{C}$.

    \Ensure List $H = [h_1,\ldots,h_n]$ such that $C =
            (t-h_1)\cdots(t-h_n)$ and $M_i = (t-h_i)(t-\cj{h_i})$ for
            $i = 1,\ldots,n$; initially $H = [\,]$ is the empty list.
    \Statex
    \Repeat
      \State $F \gets (M_1,\ldots,M_{n-1})$
      \State $R \gets \Call{Remainder}{C, M_n}$ \label{li:remainder}
      \State $h \gets \text{unique zero of $R$}$ \label{li:zero}
      \State $H \gets [h, H]$ \Comment prepend $h$ to list
      \State $C \gets \Call{Quotient}{C, t-h_i}$ \label{li:quotient}
    \Until{$\deg C = 0$}
    \State \Return $H$.
  \end{algorithmic}
\end{algorithm}

Even if most examples of motion factorization in literature are
symbolic, it is important to remark that both Algorithm~\ref{alg:qr}
and Algorithm~\ref{alg:factorization} can easily be implemented with
floating point numbers. From a numerical point of view, the most
challenging step in the computation is the factorization of
the norm polynomial $C\cj{C} = M_1 \cdots M_n$. But this is a standard
problem of numerical mathematics with a variety of fast and stable
algorithms available.

\subsection{Synthesis of Bennett linkages}
\vskip -0.2cm

The most important application of motion factorization is linkage
synthesis. As one particular example, we present the synthesis of
spatial four-bar linkages (Bennett linkages). We are not content with
the generic case alone but also discuss special cases that might
occur. Synthesis algorithms for Bennett linkages are known for a
while, see \cite{brunnthaler05:_bennet_synthesis} or \cite{perez03}
and the references therein. The approach via motion factorization
gives additional insights and, more importantly, allows
generalizations as in \cite{hegedus14:_four_pose_synthesis}.

We wish to construct a Bennett linkage that visits three given poses
$p_0$, $p_1$, $p_2$ which we view as points in $\SQ \setminus
\EG$. A crucial observation of \cite{brunnthaler05:_bennet_synthesis} is
that the coupler motion of a Bennett linkage is parameterised by a
quadratic motion polynomial that can easily be computed. Lets consider
the example
\begin{equation*}
  \begin{aligned}
    p_2 &= 1 - \qi - \qj - \qk + \eps (1 + \qk),\\
    p_1 &= 3 - \qi - 2\qj - \qk + \eps(1 - \qi + \qj + 2\qk),\\
    p_2 &= 1.
  \end{aligned}
\end{equation*}
Setting $C = t^2 + (\lambda p_1 - 1 - \mu p_0)t + \mu p_0$ with yet
undetermined $\lambda$, $\mu \in \R \setminus \{0\}$ we ensure
$C(0) = \mu p_0 \peq p_0$, $C(1) \lambda p_1 \peq p_1$,
$C(\infty) \peq p_2$. Here, the symbol ``$\peq$'' denotes equality in
projective sense, that is, up to scalar factors. Now we have to select
$\lambda$ and $\mu$ such that the dual part of the norm polynomial
$C\cj{C}$ vanishes. Comparing coefficients, we obtain the equations
\begin{equation*}
  \lambda - \mu = 2\mu - \lambda\mu - \lambda = \mu(\lambda - 1) = 0
\end{equation*}
with the unique solution $\lambda = \mu = 1$. Thus, the coupler motion
of the sought Bennett mechanism is given by the quadratic motion
polynomial
\begin{equation*}
  C = t^2 + (1-\qj)t + 1 - \qi - \qj - \qk -
      \eps((\qi - \qj - \qk)t - 1 - \qk).
\end{equation*}
It is rather typical for this approach to determine the coupler motion
before the dimension and position of the actual mechanism (``mechanism
from motion''). This can be done by computing the two factorizations
of $C$. In order to use Algorithm~\ref{alg:factorization}, we need the
quadratic, irreducible factors $M_1$ and $M_2$ of $C\cj{C}$. In our
example, we obtain
\begin{equation*}
  M_1 = t^2+2,\quad
  M_2 = t^2+2t+2.
\end{equation*}
Running Algorithm~\ref{alg:factorization} twice with $F = (M_1,M_2)$
and $F = (M_2,M_1)$ yields the two factorizations
$C = (t-h_1)(t-h_2) = (t-k_1)(t-k_2)$ where
\begin{equation}
  \label{eq:5}
  \begin{aligned}
    h_1 &= \qj + \qk + \eps(\qi + \qj-\qk), \quad& h_2 &= -1 - \qk - 2\eps\qj, \\
    k_1 &= -1-\qi -\eps\qk, \quad& k_2 &= \qi + \qj + \eps(\qi - \qj).
  \end{aligned}
\end{equation}
The quaternions $h_2$ and $k_2$ describe rotations about the moving
axes in the reference configuration $C(\infty) = 1$ (the
identity). The quaternions $h_1$ and $k_1$ give the position of the
fixed axes in that configuration. The linkage's
Denavit-Hartenberg parameters can easily be computed from the Plücker
coordinates of the moving and fixed revolute axes. The loop closure
equation is
\begin{equation*}
  \begin{aligned}
    1 &= (t - h_1) (t - h_2) (t - k_2)^{-1} (t - k_1)^{-1}\\
      &\peq (t - h_1) (t - h_2) (t - \cj{k_2}) (t - \cj{k_1})\\
      &= (t^2+2)(t^2+2t+2).
  \end{aligned}
\end{equation*}
where ``$\peq$'' denotes equality up to real polynomial factors.

We conclude this section with a hint to
\cite{hegedus14:_four_pose_synthesis}. This paper presents a synthesis
method for over-constrained 6R linkages to four prescribed poses. It
is based on rational cubic interpolation on the \emph{Study quadric}
$\SQ$ and subsequent motion factorization and can be viewed as direct
generalization of the synthesis of Bennett linkages we just
presented. The interpolation procedure is, however, more involved:
Interpolants exist only if the three-space spanned by the given poses
intersects the Study quadric $\SQ$ in a ruled quadric (a hyperboloid,
not an ellipsoid). Then, two one-parametric families of rational
interpolating cubic curves exist, each giving rise, via motion
factorization, to families of solution linkages.

\subsection{What Can Go Wrong?}
\vskip -0.2cm

Above example presented an ideal case for synthesis of Bennett
linkages. But things can go wrong in special situations. Usually,
literature on synthesis of Bennett linkages ignores these exceptional
cases but a robust synthesis procedure should be aware of them.

To begin with, the curve parameterized by $C$ may actually lie on a
straight line. In this case, the end-effector motion is either a
rotation about a fixed axis or a straight line. Situations like this
are easy to detect and we exclude them in the following. Should $C$
parameterise a planar or spherical motion, the factorization algorithm
will work and yield a planar or spherical anti-parallelogram
linkage. Note, however, that in these cases a two-parametric set of
quadratic interpolants $C$ exists, because the dual part of $C\cj{C}$
vanishes identically in $\lambda$ and~$\mu$.

The conditions for using Algorithm~\ref{alg:factorization} may not be
met, that is, the primal part $P$ of $C = P + \eps Q$ may have a real
factor. If $P$ is even real, the motion is a curvilinear translation
along a rational curve of degree two. We do not intend to give a full
discussion here. In Section~\ref{sec:exceptional-factorizations} we
will deal in detail with the cases of elliptic and circular
translations. If $P$ has precisely one real factor of degree one,
there exists a parameter value $t_0$ with $P(t_0) = 0$ and hence
$C(t_0) = \eps Q(t_0)$. Should $Q(t_0)$ be zero, the curve lies on a
straight line. Hence $Q(t_0) \neq 0$ and $C(t_0)$ lies in the
exceptional generator of the \emph{Study quadric} $\SQ$. This case has
been discussed in \cite{hamann11}. It leads to the coupler motion of
RPRP linkages \cite{perez11} and is still amenable to
Algorithm~\ref{alg:factorization} --- even if not all prerequisites
are met. An example is $C = (t - h_1)(t-h_2) = (t - k_1)(t - k_2)$
where
\begin{alignat*}{2}
 h_1 &= \qi + \qj + \eps(\qi - \qj), \quad& h_2 &= 1 + \eps(\qi + 3\qj - \qk),\\
  k_1 &= 1 + \eps(3\qi + \qj - \qk), \quad& k_2 &= \qi + \qj - \eps(\qi - \qj).
\end{alignat*}
Note that $h_1$ and $k_2$ are rotation quaternions with (necessarily)
parallel axes while $h_2$ and $k_1$ are translation quaternions.

Finally, we also would like to discuss a situation that was pointed
out by M.~Hamann (Dresden University of Applied Sciences). It is
possible that Algorithm~\ref{alg:factorization} is applicable but does
not yield a proper linkage because the two synthesized 2R dyads
\emph{coincide.}  Using motion factorization, we can easily
characterize this situation and construct a replacement mechanism.

It is known \cite{krames37:_bennett_mechanismus} that the coupler
motion of Bennett linkages is obtained by reflecting a fixed system in
the rulings of one family on a hyperboloid of revolution (it is
\emph{line symmetric} with respect to that system of rulings). Using
the hyperboloid with equation
$x^2b^2c^2 + y^2a^2c^2 - z^2a^2b^2 = a^2b^2c^2$ and
$a,b,c \in \R \setminus \{0\}$, a possible motion polynomial reads
\begin{equation}
  \label{eq:6}
  \begin{aligned}
    C &= (b^2+c^2)t^2 \\
      &+ (2a(c\qj + b\qk) - 2\eps(a^2(b\qj - c\qk) - bc(b\qk+c\qj)))t \\
      &+ 2bc\qi - b^2 + c^2 - 2a\eps((b^2-c^2)\qi + 2bc)
  \end{aligned}
\end{equation}
Two factorizations
$C = (b^2+c^2)(t-h_1)(t-h_2) = (b^2+c^2)(t-k_1)(t-k_2)$ can only
coincide if the norm polynomial factors as $C\cj{C} = M^2$ with a
quadratic, irreducible polynomial $M \in \R[t]$. From the discriminant
\begin{equation}
  \label{eq:7}
  4096(b^2+c^2)^8(a^2+c^2)^2(a+b)^2(a-b)^2
\end{equation}
of $C\cj{C}$ we read off that this can only happen if $a = b$. Indeed, we
then have $C\cj{C} = (a^2+c^2)^2(t^2+1)^2$. Thus, we have

\begin{proposition}
  A line symmetric motion with respect to one family of lines on a
  hyperboloid $H$ can be generated by only one 2R dyad if and only if $H$
  is a hyperboloid of revolution. Otherwise, there exist two 2R dyads.
\end{proposition}

In other words, synthesis of a Bennett linkage fails if the three
input poses are sampled from a line symmetric motion with respect
to a hyperboloid of revolution.

\subsection{Bennett flips}
\vskip -0.2cm
\label{sec:bennett-flips}

\begin{figure}
  \centering
  \includegraphics{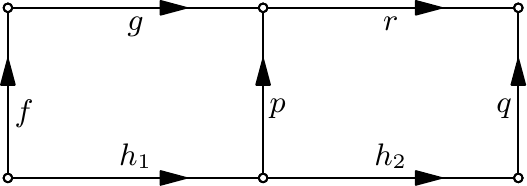}
  \caption{Link graph for construction of replacement mechanisms}
  \label{fig:diagram}
\end{figure}

Motion factorization not only reveals a caveat of existing Bennett
synthesis methods but can also be used for the construction of a
replacement linkage with five or six joints. This construction is
based on an important technique for rational linkages which we call
\emph{Bennett flip.} Consider the motion polynomial
$C = (t-h_1)(t-h_2)$, pick an arbitrary rotation quaternion $p$ and
use Algorithm~\ref{alg:factorization} to compute rotation quaternions
$f$, $g$, $r$, $q$ such that
\begin{gather*}
  (t-h_1)(t-p) = (t-f)(t-g),\\
  (t-\cj{h_2})(t-p) = (t-q)(t-\cj{r}).
\end{gather*}
The link diagram of this construction is depicted in
Figure~\ref{fig:diagram}. The vertices indicate links and the arrows
with label $x$ connect links such that the relative motion of the
second (at the arrow tip) with respect to the first (at the arrow
base) is parameterised by $t-x$.  We have the loop closure equations
\begin{equation}
  \label{eq:8}
  \begin{aligned}
    1 &\peq (t-h_1)(t-p)(t-\cj{g})(t-\cj{f}) \\
      &\peq (t-q)(t-\cj{r})(t-\cj{p})(t-h_2).
  \end{aligned}
\end{equation}
The second closure equation implies
$(t-q)(t-\cj{r}) \peq (t-\cj{h_2})(t-p)$ so that
\begin{equation}
  \label{eq:9}
  \begin{aligned}
  &\phantom{\;\peq\;} (t-h_1)(t-h_2)(t-q)(t-\cj{r})(t-\cj{g})(t-\cj{f}) \\
  &\peq (t-h_1)(t-h_2)(t-\cj{h_2})(t-p)(t-\cj{g})(t-\cj{f}) \\
  &\peq (t-h_1)(t-p)(t-\cj{g})(t-\cj{f}) \\
  &\peq 1.
  \end{aligned}
\end{equation}
The last equality follows from the first closure equation in
\eqref{eq:8}. From \eqref{eq:9} we infer that the rotation quaternions
$h_1$, $h_2$, $q$, $\cj{r}$, $\cj{g}$, $\cj{f}$ can be assembled, in
that order, to form an over-constrained 6R linkage. As long as $p$ is
chosen sufficiently general, this construction works and the resulting
6R linkage (of type Waldron's double Bennett hybrid) has precisely one
degree of freedom. The link connecting $h_2$ and $q$ performs the
motion parameterized by $C$.  Note that this construction does not
hinge on the vanishing of the discriminant \eqref{eq:7} but is valid
for arbitrary Bennett motions.

A special choice of $p$ may even cause the axis of $g$ and $r$ to
coincide. This then yields an over-constrained 5R linkage (a Goldberg
mechanism), for which $(t-h_1)(t-h_2)$ parameterises the motion of one
link. Let's look at the concrete example obtained by plugging $a = b =
1$ and $c = 2$ into the motion polynomial \eqref{eq:6}. It has the
factorization $C = (t-h_1)(t-h_2)$ where
\begin{equation*}
  h_1 = -\tfrac{1}{5}(4\qj - 3\qk + 2\eps(3\qj + 4\qk)),\quad
  h_2 = -\qk.
\end{equation*}
Setting $p = \qj - \qk + \tfrac{5}{2}\eps(\qj + \qk)$ we compute the
rotation quaternions $f$, $g$, $r$, and $q$ via Bennett flips:
\begin{equation*}
  \begin{aligned}
    f &= \tfrac{1}{5}(\qj - 7\qk) + \tfrac{9}{10}\eps(7\qj + \qk),\\
    g &= \qk - 5\eps\qj = r,\\
    q &= \qj + \qk - \tfrac{5}{2}\eps(\qj - \qk).
  \end{aligned}
\end{equation*}
Because $g = r$, two consecutive axes coincide and the corresponding
linkage has only five revolute joints.

\section{Exceptional Factorizations}
\label{sec:exceptional-factorizations}
\vskip -0.2cm

Motion polynomials whose primal part has no real factor can always be
factorized by Algorithm~\ref{alg:factorization}. However, many
rational motions of practical and theoretical importance are not
generic. For those motions, computation of the unique zero $h_i$ of a
linear remainder polynomial in Line~\ref{li:zero} of
Algorithm~\ref{alg:factorization} may fail. Note that motion
polynomials in $\H[t]$ are always projectively equal to generic motion
polynomials and Algorithm~\ref{alg:factorization} always works. This
is an example for the failure of naive transfer of results of
spherical kinematics to spatial kinematics via wrong application of
Kotelnikov's and Study's principle of transfer.

Failure of Algorithm~\ref{alg:factorization} does not imply anything
on existence or non-existence of factorizations. There exist rational
motions with no factorization and rational motions with infinitely
many factorization.

\subsection{Elliptic translations}
\label{sec:elliptic-translations}
\vskip -0.2cm

Let us discuss, as an example, the motion polynomial
\begin{equation}
  \label{eq:10}
  C = t^2 + 1 + \eps (b\qj t + a\qi),
  \quad
  a \ge b > 0.
\end{equation}
It parameterizes an elliptic translation. The primal part of $C$ is
real so that Algorithm~\ref{alg:factorization} is not
applicable. But we can try to manually solve the equation
\begin{equation}
  \label{eq:11}
  C = (t-h_1)(t-h_2)
\end{equation}
for $h_1 = p_1 + \eps d_1$, $h_2 = p_2 + \eps d_2$. Comparing
coefficients of the primal part, we immediately find the conditions
$p_1 + p_2 = 0$, $p_1p_2 = 1$ and the solution
\begin{equation*}
  p_1 = x\qi + y\qj + z\qk,\
  p_2 = \cj{p_1}
  \quad\text{where}\quad
  x^2 + y^2 + z^2 = 1.
\end{equation*}
The dual parts are $d_i = u_i\qi + v_i\qj + w_i\qk$ for $i =
1,2$. Comparing coefficients gives the additional equations
\begin{equation}
  \label{eq:12}
  \begin{aligned}
    (u_1 - u_2) x + (v_1 - v_2) y + (w_1 - w_2) z &= 0 \\
    (v_1 + v_2) z - (w_1 + w_2) y + a &= 0 \\
    (u_1 + u_2) z - (w_1 + w_2) x &= 0 \\
    (u_1 + u_2) y - (v_1 + v_2) x &= 0 \\
    u_1 + u_2 = b + v_1 + v_2 = w_1 + w_2 &= 0.
  \end{aligned}
\end{equation}
Th system \eqref{eq:12} has the solution
\begin{gather*}
  x = 0,\ z = \frac{a}{b},\
  u_1 = \lambda,\
  u_2 = -\lambda,\
  v_1 = \mu - b,\
  v_2 = -\mu,\\
  w_1 = \frac{yb(b-2\mu)}{2a},\
  w_2 = -\frac{yb(b-2\mu)}{2a},
\end{gather*}
with $\lambda$, $\mu \in \R$. From $x^2 + y^2 + z^2 = 1$ and
$a \ge b > 0$ we infer $a = b$ and $y = 0$ whence the solution becomes
\begin{gather*}
  x = y = 0,\ z = 1,\ u_1 = \lambda,\ u_2 = -\lambda,\\
  v_1 = \mu - a,\ v_2 = -\mu,\ w_1 = 0,\ w_2 = 0.
\end{gather*}

In summary, we have shown:

\begin{proposition}
  Elliptic translations do not admit factorizations of the shape
  $(t-h_1)(t-h_2)$ but circular translations admit infinitely many. If
  the circular translation is parameterized as in \eqref{eq:10} with
  $a = b$, these factorizations are given by
  $h_1 = \qk + \eps(\lambda\qi + (\mu - a)\qj)$,
  $h_2 = -\qk - \eps(\lambda\qi + \mu\qj)$.
\end{proposition}

\begin{figure}
  \centering
  \begin{overpic}{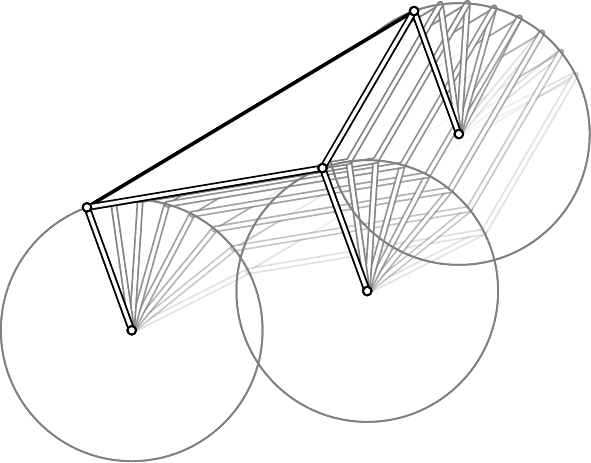}
    \footnotesize
    \put(16,20){$h_1$}
    \put(9,46){$h_2$}
    \put(64,24){$k_1$}
    \put(50,53){$k_2$}
    \put(79,53){\contourlength{1pt}\contour{white}{$\ell_1$}}
    \put(65,79){$\ell_2$}
  \end{overpic}
  \caption{Three different factorizations of a circular translation}
  \label{fig:parallelogram}
\end{figure}
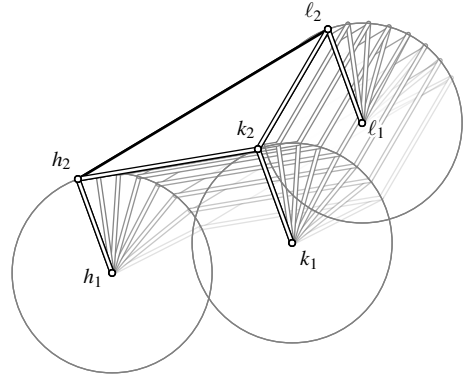

In fact, it is well-known that a circular translation occurs in
infinitely many ways as coupler motion of a parallelogram
linkage. Each factorization corresponds to the motion of one 2R chain
in such a linkage. This is illustrated in
Figure~\ref{fig:parallelogram}.

\subsection{The Multiplication Trick}
\label{sec:multiplication-trick}
\vskip -0.2cm

The elliptic translation \eqref{eq:10} admits no factorization with two
linear motion polynomials. But there is a way to obtain factorizations
with four motion polynomials. The basic idea is to find a real,
irreducible polynomial $Q$ such that $QC$ admits a factorization. Note
that $C$ and $QC$ are equal in projective sense, that is, they
parameterise the same motion. By a result of
\cite{li15:_motion_polynomials}, this is always possible and the
degree of the real polynomial $Q$ is bounded by the degree of the
greatest real factor in the primal part of~$C$.

\begin{figure}
  \centering
    \includegraphics{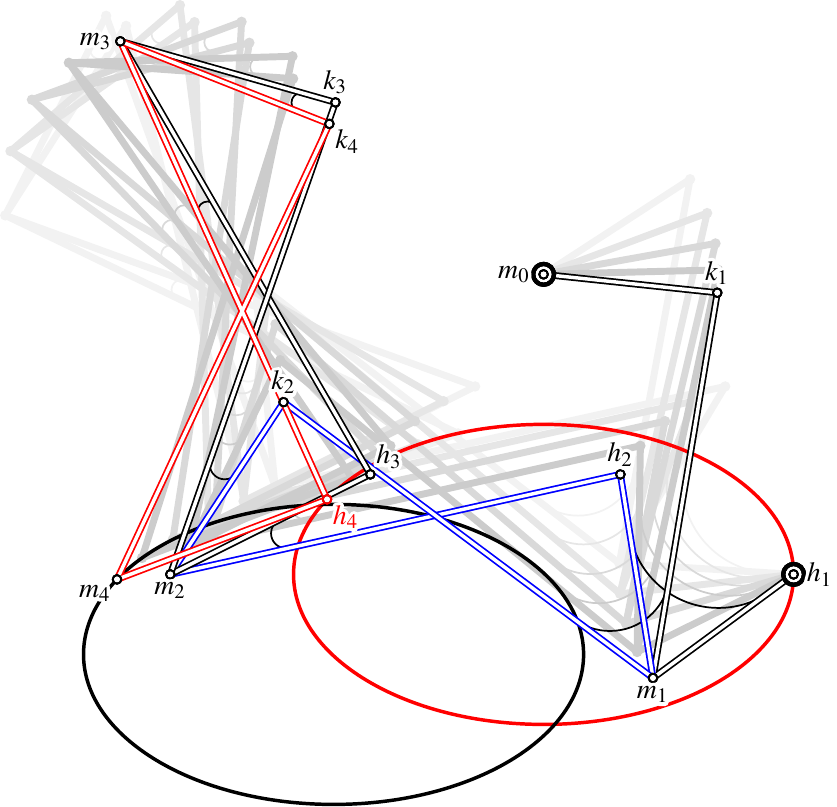}\\[1.5ex]
  \includegraphics{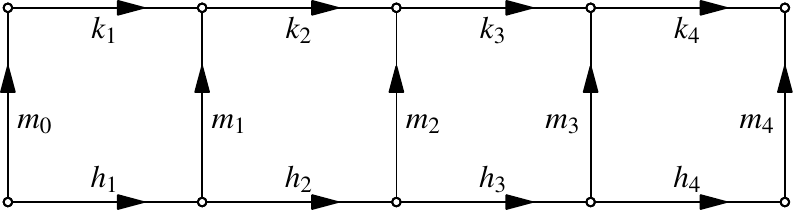}
  \caption{Linkage to generate an elliptic translation and
    corresponding link graph}
  \label{fig:elliptic-translation}
\end{figure}

Returning to the elliptic translation \eqref{eq:10} and look at one
particular example. Assuming $a \neq b$, we let $Q = t^2 + 1$ and compute
$QC = (t-h_1)(t-h_2)(t-h_3)(t-h_4)$ where
\begin{equation}
  \label{eq:13}
  \begin{aligned}
    h_1 &= \phantom{-}\qk - \frac{a-b}{a+b}\eps\qi,\\
    h_2 &= -\qk + \frac{(2(a-b)\qi - (a+b)^2\qj)\eps}{2(a+b)},\\
    h_3 &= -\qk - (\qi - \tfrac{1}{2}(a-b)\qj)\eps,\\
    h_4 &= \phantom{-}\qk + \qi\eps.
  \end{aligned}
\end{equation}
By a construction similar to the Bennett flip in
Section~\ref{sec:bennett-flips}, we may ``brace'' each link $h_i$ with
an anti-parallelogram linkage with joints $h_i$, $m_i$, $k_i$
$m_{i-1}$. Combing these anti-parallelograms gives a linkage where the
link connecting $m_4$ and $h_4$ performs an elliptic translation
(Figure~\ref{fig:elliptic-translation}). The construction is
initialized by picking a generic planar rotation quaternion $m_0$ and
recursively computing $k_i$ and $m_i$ from
\begin{equation*}
  (t - m_{i-1}^{-1})(t-h_i) = (t - k_i)(t - m_i^{-1}),
  \quad i = 1,2,3,4.
\end{equation*}
The joints triplets $h_i$, $m_i$, $h_{i+1}$ and $k_i$, $m_i$,
$k_{i+1}$, respectively, belong to the same link. This is indicated by
circular arcs in Figure~\ref{fig:elliptic-translation}. Constructions
of this type are at the core of \cite{gallet15} where bounded rational
curves of arbitrary degree are drawn by planar linkages.

The factorization \eqref{eq:13} is by no means unique. We also have
the strange factorization $QC = (t-k_1)(t-k_2)(t-k_3)(t-k_4)$ where
\begin{equation*}
  \begin{aligned}
    k_1 &= -\frac{(a^2-b^2)\qi-2ab\qk}{a^2+b^2} - \frac{(a^2-b^2)\eps\qj}{a^2+b^2},\\
    k_2 &= \frac{(a^2-b^2)\qi - 2ab\qk}{a^2+b^2} - \frac{((a^2+b^2)^2-2b(a^2-b^2))\eps\qj}{2b(a^2+b^2)},\\
    k_3 &= -\qi + \frac{a^2-b^2+2b}{2b}\eps\qj,\\
    k_4 &= \qi - \eps\qj.
  \end{aligned}
\end{equation*}
In contrast to \eqref{eq:13}, this factorization is \emph{not planar.}
It gives rise to a spatial open chain with parallel first and second
and parallel third and fourth axis that can follow a curvilinear
translation along an elliptic path. We may combine the four revolute
axes of the latter factorization with two prismatic joints to obtain
an an over-constrained RRRRPP mechanism one of whose links performs an
elliptic translation.

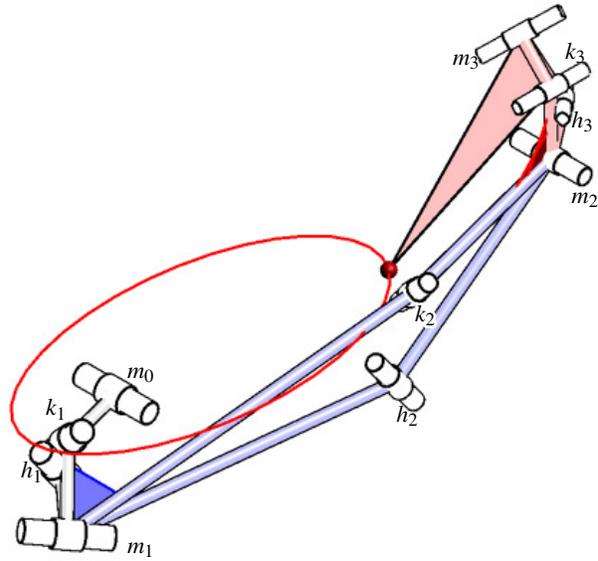
\begin{figure}
  \centering
  \begin{overpic}[trim=53 80 43 23,clip,width=\linewidth]{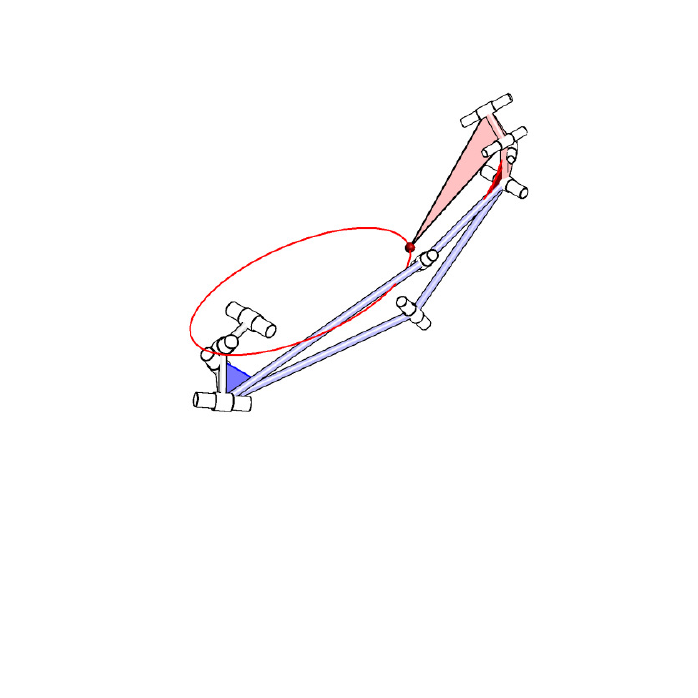}
    \small
    \put(20,30){$m_0$}
    \put(3,14){$h_1$}
    \put(7,24){$k_1$}
    \put(20,2){$m_1$}
    \contourlength{1pt}
    \put(66,39){\contour{white}{$k_2$}}
    \put(91,58){$m_2$}
    \put(63,23){$h_2$}
    \put(90,81){$k_3$}
    \put(72,80){$m_3$}
    \put(91,70){$h_3$}
  \end{overpic}
  \caption{Spatial linkage to generate a Darboux motion.}
  \label{fig:darboux}
\end{figure}

If we abandon the requirement to generate an elliptic translation, we
may also make the motion polynomial $C$ of \eqref{eq:10} factorizable
by right-multiplying it with a suitable quaternion polynomial
$H \in \H[t]$. This changes the motion but not the trajectory of the
origin in the moving coordinate frame. If $H$ is linear, the motion
polynomial $CH$ parameterizes a Darboux motion (a spatial motion with
only planar trajectories) and our Bennett flip construction produces a
spatial linkage to draw an ellipse (Figure~\ref{fig:darboux}). The
anti-parallelograms of Figure~\ref{fig:elliptic-translation} are
replaced by Bennett linkages and only ten joints and eight links are
required. While the linkage has theoretical full-cycle mobility, it
is, of course, a demanding engineering task to realize this in a
practical design. For more information about Darboux linkages we refer
to \cite{li15:_7R}. There, also considerable refinements of the above
construction can be found.

\section{Conclusions}
\label{sec:conclusions}
\vskip -0.2cm

This article gives an overview of algorithms and techniques associated
with the factorization of motion polynomials. It addresses several
special but yet instructive cases and gives concrete applications. We
complemented known methods for the synthesis of Bennett linkages by a
discussion of a special case that so far has skipped attention of
researchers and we showed how to use Bennett flips to produce
replacement linkages. We also hinted at a new method for the
construction of rational over-constrained 6R linkages to four
prescribed poses. The attentive reader may have noticed a gap in the
chain of synthesis procedures: We did not talk about 5R linkages. In
fact, it is possible to synthesis 5R linkages to three poses plus
``some additional constraints''. An efficient algorithm to do this
hinges on a better understanding of three-spaces generated by 2R
dyads and will be addressed in future work.

In a second part, we gave a preview on very recent methods to factor
exceptional motion polynomials that evade the algorithms for generic
situations. The ``multiplication trick'' is actually a powerful tool
to factor all motion polynomials and, together with Bennett flips,
belongs to a toolbox for the construction of linkages to arbitrary
rational motions or to arbitrary rational trajectories.  The general
theory and algorithmic treatment of the ``multiplication trick'' is
currently being worked out. As these theoretical constructions do not
necessarily require an excessive number of links, there is hope for
useful engineering applications.

\section*{Acknowledgments}
\vskip -0.2cm

This research was supported by the Austrian Science Fund (FWF):
P\;26607 (Algebraic Methods in Kinematics: Motion Factorization and
Bond Theory).

\bibliographystyle{iftomm}

\end{document}